\documentclass[letterpaper]{article} % DO NOT CHANGE THIS
\usepackage{aaai25}  % DO NOT CHANGE THIS
\usepackage{times}  % DO NOT CHANGE THIS
\usepackage{helvet}  % DO NOT CHANGE THIS
\usepackage{courier}  % DO NOT CHANGE THIS
\usepackage[hyphens]{url}  % DO NOT CHANGE THIS
\usepackage{graphicx} % DO NOT CHANGE THIS
\usepackage{natbib}  % DO NOT CHANGE THIS AND DO NOT ADD ANY OPTIONS TO IT
\usepackage{caption} % DO NOT CHANGE THIS AND DO NOT ADD ANY OPTIONS TO IT
\usepackage{algorithm}
\usepackage{algorithmic}
\usepackage{newfloat}
\usepackage{listings}
\usepackage{amssymb}
\usepackage{amsmath}
\usepackage{booktabs} 
\usepackage{multirow} 
\usepackage{marvosym}
\makeatletter
\def\@fnsymbol#1{\ensuremath{\ifcase#1\or \textsuperscript{\Letter}\or \ddagger\or
   \mathsection\or \mathparagraph\or \|\or **\or \dagger\dagger
   \or \ddagger\ddagger \else\@ctrerr\fi}}
\makeatother

\DeclareCaptionStyle{ruled}{labelfont=normalfont,labelsep=colon,strut=off} % DO NOT CHANGE THIS
\floatstyle{ruled}
\newfloat{listing}{tb}{lst}{}
\floatname{listing}{Listing}
\title{Attend and Enrich: Enhanced Visual Prompt for Zero-Shot Learning}
\author{
Man Liu$^{2,4}$, Huihui Bai$^{1,2,4}$\textsuperscript{\Letter}, Feng Li$^{3}$\textsuperscript{\Letter},
  Chunjie Zhang$^{2,4}$,\\
  Yunchao Wei$^{2,4}$, Tat-Seng Chua$^{5}$,
  Yao Zhao$^{2,4}$
}
\affiliations{
    \textsuperscript{\rm 1} Tangshan Research Institute of Beijing Jiaotong University  \\
    \textsuperscript{\rm 2}Institute of Information Science, Beijing Jiaotong University \\
    \textsuperscript{\rm 3} Hefei University of Technology  \\
    \textsuperscript{\rm 4} Beijing Key Laboratory of Advanced Information Science and Network Technology 
     \\
  \textsuperscript{\rm 5} National University of Singapore\\
 \{manliu, hhbai, cjzhang, yunchao.wei, yzhao\}@bjtu.edu.cn, 
  fengli@hfut.edu.cn, chuats@comp.nus.edu.sg
  \phantom{\thanks{Corresponding author.}}
}
\usepackage{bibentry}
\begin{document}

\maketitle

\begin{abstract}
Zero-shot learning (ZSL) endeavors to transfer knowledge from the seen categories to recognize unseen categories, which mostly relies on the semantic-visual interactions between image and attribute tokens. Recently, the prompt learning has emerged in ZSL and demonstrated significant potential as it allows the zero-shot transfer of diverse visual concepts to downstream tasks. However, current methods explore the fixed adaptation of the learnable prompt on the seen domains, which make them over-emphasize the primary visual features observed during training, limiting their generalization capabilities to the unseen domains. In this work, we propose AENet, which endows semantic information into the visual prompt to distill semantic-enhanced prompt for visual representation enrichment, enabling effective knowledge transfer for ZSL. AENet comprises two key steps: 1) exploring the concept-harmonized tokens for the visual and attribute modalities, grounded on the modal-sharing token that represents consistent visual-semantic concepts; and 2) yielding the semantic-enhanced prompt via the visual residual refinement unit with attribute consistency supervision. It is further integrated with primary visual features to attend to semantic-related information for visual enhancement, thus strengthening transferable ability. Experimental results on three benchmarks show that our AENet outperforms existing state-of-the-art ZSL methods.
\end{abstract}
 \begin{links}
     \link{Code}{https://github.com/ManLiuCoder/AENet.}
 %    \link{Extended version}{https://aaai.org/example/extended-version}
\end{links}
\section{Introduction}
\begin{figure}[t]
\begin{center}
\includegraphics[scale=0.45]{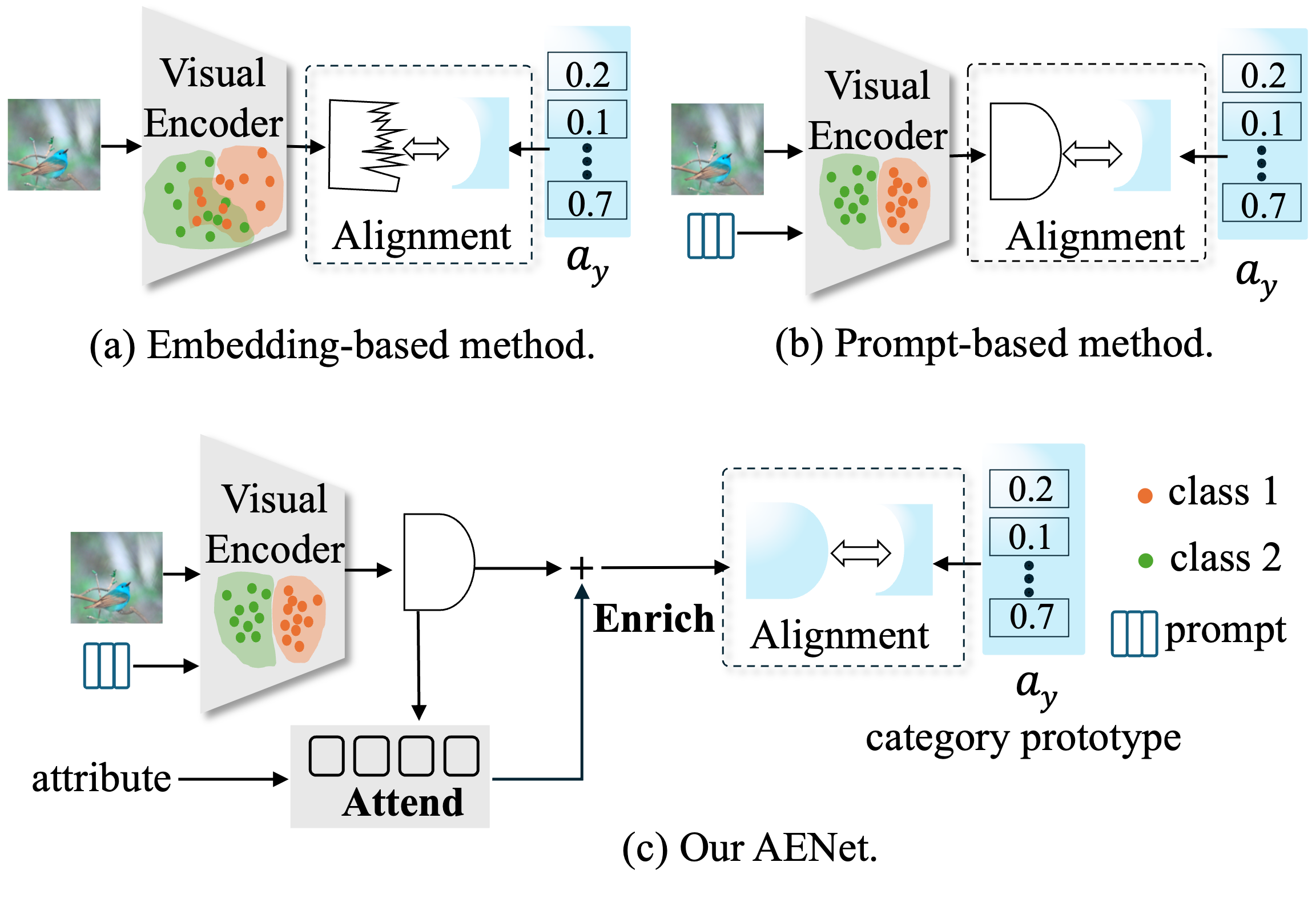}
\caption{\textbf{Motivation of AENet.} (a) Embedding-based methods align visual and semantic features for ZSL, which suffer from vague visual representations due to cross-dataset bias. (b) Prompt-based methods add the learnable prompt to pre-trained vision encoders, adapting them to ZSL scenarios while over-emphasizing primary visual features. (c) Our AENet distills the semantic-enhanced prompt that attends to semantic-related details to enrich visual representations, ensuring more comprehensive knowledge transfer.}

\label{fig:motivation}
\end{center}
\end{figure}

Humans naturally interact with the world through various channels such as vision and language, enabling them to identify unseen objects based on prior knowledge. Leveraging this cognitive capability, zero-shot learning (ZSL) \cite{palatucci2009zero,Lampert2009LearningTD} aims to classify objects from the unseen domain carrying knowledge from seen categories. One of the mainstream methods in ZSL is the embedding-based approach \cite{APN2020,GEM2021,MSDN2022,TransZero,POPRNet}, which learns the semantic-visual alignment in a joint embedding space. Prior to such a cross-modal alignment, as depicted in Figure \ref{fig:motivation}(a), embedding-based models typically commence with a visual encoder, such as ResNet~\cite{Resnet2016} or ViT~\cite{ViT2020} pre-trained on ImageNet \cite{ImageNet}, to initialize visual features. However, this can introduce distribution discrepancy when applied to downstream ZSL benchmarks as wider categories are unobserved during pre-training, thereby resulting in cross-dataset bias that produces vague or even misleading visual representations~\cite{2011Unbiased}.

To mitigate the distribution discrepancy and cross-dataset bias issues arising from using pre-trained encoders, recent ZSL methods have incorporated attention mechanisms~\cite{AREN2019,SGMA2019} and vision transformers~\cite{TransZero,TransZero++,ZSLViT}. These approaches refine visual representations by focusing on informative image regions and capturing complex visual-semantic relationships. Recently, motivated by the powerful zero-shot capabilities of prompt engineering \cite{kojima2022large,cheng2023uprise}, some works \cite{VPT2022,coop,cocoop,ship} propose to incorporate the learnable visual prompt into pre-trained vision encoders to adapt the visual space to downstream datasets (Figure~\ref{fig:motivation}(b)), which effectively alleviates the cross-dataset bias. However, in these methods, the learned prompt often tends to overfit the base seen classes and concentrate solely on primary visual features essential for recognizing seen categories~\cite{cocoop}. Consequently, they may lack sufficient capacity to capture crucial semantic-related visual features necessitated for a broader range of unseen classes. These features, which are complementary to primary contents, are pivotal for effective cross-domain semantic transfer.

To address these issues, in this work, we propose a novel method, which embraces prompt learning with ZSL. This method follows the existing prompt-based pipeline but investigates concept-harmonized semantic features to provide the semantic-enhanced prompt for visual enhancement, thereby achieving more comprehensive knowledge discovery. It takes the attribute, image, and initialized learnable visual prompt as input, to obtain corresponding embeddings. Given the inherent disparity between the image and attribute, we first devise the concept-aware attention (CAA) to harmonize the concept of visual and attribute tokens. This attention defines a modal-sharing token as the reference that adapts both modalities to concept-harmonized tokens, which are expected to reveal consistent semantic and visual concepts. Then, a visual residual refinement unit (VRRU) operating in a simple yet efficient linear manner is constructed to mine semantic-related details under attribute consistency supervision. In this way, we can form the semantic-enhanced prompt that attends to semantic-related features by combining the predicted residuals with the initial prompt embedding. They are further utilized to enrich the primary visual representations, facilitating the efficacy of visual-semantic alignment for unseen domain recognition. We call this method AENet. Extensive experimental results show that the proposed AENet leads to better performance on ZSL benchmarks.

In summary, our work makes the following contributions: 1) We propose AENet that leverages prompt learning with concept-harmonized semantic features to enhance visual-semantic alignment for ZSL. 2) We introduce the concept-aware attention to harmonize visual and attribute concepts grounded on the modal-sharing token. 3) We propose VRRU, a linear unit which mines semantic-related details under attribute consistency supervision, generating the semantic-enhanced prompt through the combination of predicted residuals.

\section{Related Work}
\subsection{Zero-Shot Learning}
ZSL approaches primarily fall into two categories: generative-based and embedding-based methods. Generative methods synthesize visual features for unseen categories using techniques like generative adversarial networks \cite{CEGZSL2021} or variational autoencoders \cite{HSVA2021,SDGZSL2021}. Although these methods compensate for the absence of the unseen domain during training, they introduce extra data. The embedding-based method represents the other mainstream branch of GZSL, achieved through the projection and alignment of information from visual and semantic modalities. Early works \cite{akata2013label,Xian2016LatentEF} propose to align images and attributes within a common feature space. 
However, global visual information falls short of capturing the subtle yet substantial differences between categories, weakening discriminative representations. Thus, part-based methods attempt to highlight the most important parts of the input, better refining the extracted visual representation. For example, some investigations \cite{APN2020,DPPN2021,MSDN2022,PSVMA2023} have pursued semantic-guided attention to localize discriminative attribute-related parts and capture crucial fine-grained features. Distinctive visual features are also emphasized by graph networks \cite{RGEN2020,hu2021graph} that incorporate region-based relational analysis to mine complementary connections among diverse spatial elements. However, they are still confused by the cross-dataset bias issues arising from pre-trained encoders. In contrast to the previous methods, we embrace the idea of prompt learning and enrich the visual features for desirable visual-semantic alignment.

\subsection{Prompt Learning}
Prompt learning emerges as a pivotal methodology within the realm of natural language processing (NLP), which can adapt Large Language Models (LLMs) to various downstream tasks and scenarios \cite{liu2023pre,10687525,tan2024c2p}. Initially, the utilization of prompt learning depends primarily on hand-crafted prompts, which are composed of thoughtfully selected words or phrases. While potentially powerful, it presents non-trivial prompt engineering challenges that require not only extensive domain knowledge but also consume significant time due to its inherently iterative nature of trial and error. Studies such as \cite{lester2021power} and \cite{liu2023gpt} attempt to introduce a learnable continuous prompt that is updated dynamically during the learning process, boosting model performance and efficiency. The success of prompt learning in LLMs has sparked interest in computer vision \cite{llava,wang2024revisiting}. CoOp \cite{coop} enhances pre-trained vision-language models for downstream image classification through continuous prompt learning. Several methods~\cite{cocoop,ship} improve CoOp by introducing image-conditioned tokens or synthesized prompts for underrepresented classes, which advance the capabilities of vision-language models in few/zero-shot learning. However, these models mainly focus on primary visual content sufficient for seen classes but ignore crucial visual details that are critical for unseen domains. In this paper, we focus on improving the prompt-based content by supplementing semantic-related information for visual enhancement, allowing it generalize well to the unseen classes.

\begin{figure*}[ht]
\begin{center}
\includegraphics[scale=0.65]{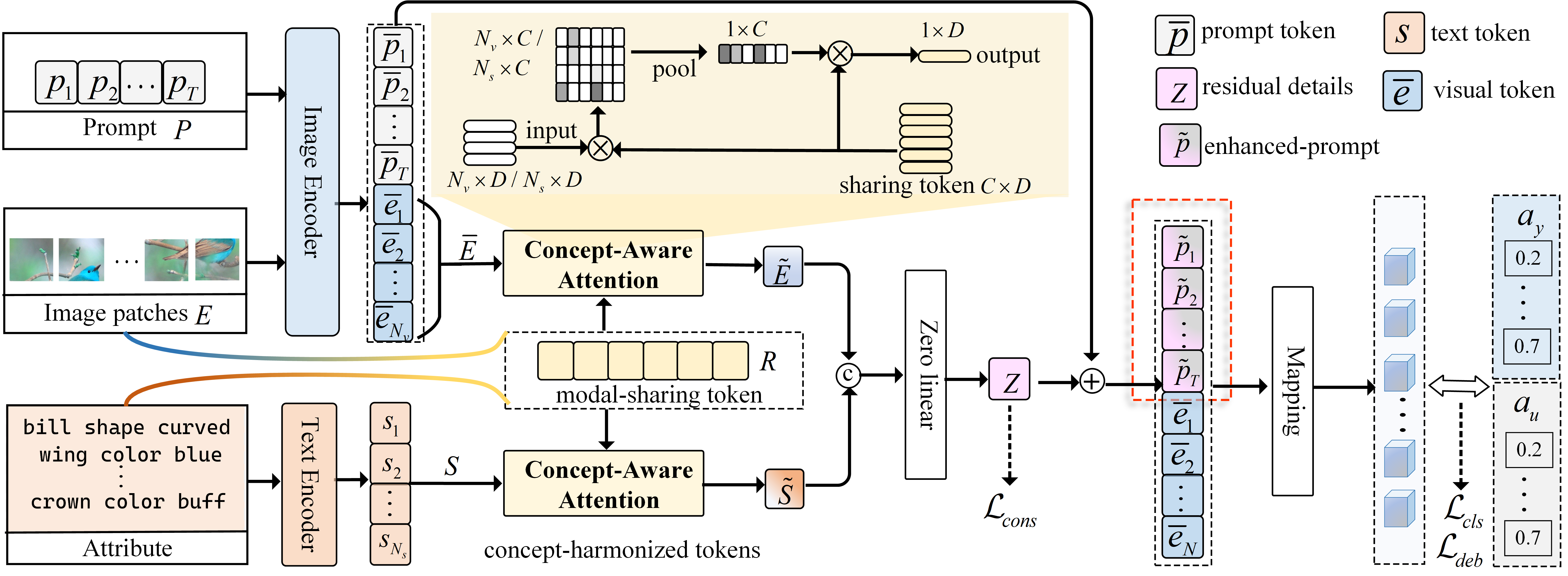}
\caption{
AENet processes attributes, images, and the learnable prompt through the pre-trained encoder. It performs concept-harmonized token exploration using concept-aware attention, followed by semantic-enhanced prompt distillation with attribute consistency. The enhanced prompt is then integrated with visual features to attend to semantic-related information, improving knowledge discovery for unseen classes.}
\label{fig:P2P}
\end{center}
\end{figure*}

\section{Methodology}
The proposed AENet aims to explore the semantic-enhanced prompt, which enriches primary visual representations to conduct accurate semantic-visual alignment for ZSL. As shown in Figure~\ref{fig:P2P}, AENet takes the attribute, image, and initialized learnable visual prompt as input to produce corresponding embeddings via the text and image encoders. Upon these, there are two key steps in AENet: 1) Concept-harmonized token exploration for visual and attribute modalities, which is grounded on the modal-sharing token that encodes consistent visual-semantic concepts by using the concept-aware attention; 2) Semantic-enhanced prompt distillation via the visual residual refinement unit with attribute consistency supervision. After that, we integrate the enhanced prompt with visual features to attend to semantic-related information for visual enhancement, achieving more effective and comprehensive knowledge discovery. 

\subsection{Problem Formulation}
%先说数据集的定义
ZSL aims to discern the novel image categories within the unseen domain $\mathcal{D}^u$ by capturing knowledge derived from the seen domain data $\mathcal{D}^s$. Here, $\mathcal{D}^s=\{(x, y,a_y )|x \in \mathcal{X}^{s},y \in \mathcal{Y}^{s},a_y \in \mathcal{A}^{s}\}$ consists of the images $x$ in $\mathcal{X}^{s} $, corresponding label $y$, and its associated category prototype $a_y$ from $\mathcal{A}^{s}$. Similarly, the unseen domain data is defined as $\mathcal{D}^u=\{(x^{u}, u,a_{u} )\}$, where $x^{u} \in \mathcal{X}^{u}$, $u \in \mathcal{Y}^{u}$, $a_{u} \in \mathcal{A}^{u}$, with $\mathcal{A}=\mathcal{A}^{s} \cup \mathcal{A}^{u}$. Note that the category space is disjoint between seen and unseen domains, \emph{i.e.}, $\mathcal{Y}^{s} \cap \mathcal{Y}^{u}=\varnothing$, $ \mathcal{Y}^{s} \cup \mathcal{Y}^{u}=\mathcal{Y}$. 

Utilizing $\mathcal{D}^s$ during the training stage, the typical embedding-based ZSL framework learns a mapping function $\mathcal{M}(\cdot)$ that bridges the image space $\mathcal{X}^{s}$ with the attribute space $\mathcal{A}^{s}$. This enables the model to generalize its knowledge to $\mathcal{D}^u$ and establish connections between $\mathcal{X}^{u}$ and $\mathcal{A}^{u}$ for category inference. In scenarios where the test phase includes both seen and unseen classes, conventional ZSL is often extended to generalized zero-shot learning (GZSL), rendering it more applicable to real-world scenarios. Given an input image representation $f(x)$ during training, the optimization of the embedding-based framework is conducted through the visual-semantic alignment, as follows:
\begin{equation}
{\mathcal{L}_{cls}} =  - \sum\limits_{x \in {X^s}} {\log \frac{{\exp \left\langle {\mathcal{M}(f(x)),{a_y}} \right\rangle }}{{\sum\limits_{\hat y \in {\mathcal{Y}^S}} {\exp } \left\langle {\mathcal{M} (f(x)),{a_{\hat y}}} \right\rangle }}} 
\label{eq:cls}
\end{equation}
where the mapping function $\mathcal{M}(\cdot)$ is generally implemented via global average pooling (GAP) and linear projection. $\left\langle\cdot\right\rangle$ represents the cosine similarity used for category decision.

\subsection{Concept-Harmonized Exploration}
As shown in Figure~\ref{fig:P2P}, taking visual image patches $E$ and the initialized visual prompt $P$ as input, inspired by~\cite{VPT2022}, we extract visual tokens $\bar{E}\in\mathbb{R}^{N_v\times D}$ from patches $E$ conditioned on $P$ using the transformer layers of ViT~\cite{ViT2020} and produce the corresponding prompt embedding $\bar{P}\in\mathbb{R}^{T\times D}$. Similarly, we can obtain the embedded sharing attribute token $S\in\mathbb{R}^{N_s\times D}$ by encoding the word vectors of each attribute descriptor with GloVe~\cite{Pennington2014GloveGV}.

\subsubsection{Concept-Aware Attention}
As illustrated in Figure \ref{fig:P2P}, the sharing attributes are incorporated with the prompt-based visual features to provide hybrid multimodal information, serving as the foundation for semantic-enhanced prompt learning. However, visual images and textual descriptions inherently differ in terms of semantic levels and granularity~\cite{shao2022learning,chen2023revisiting}. Without explicit visual-semantic alignment, the attribute-related features are prone to sub-optimal representation. Thus, the concept-aware attention (CAA) is proposed to harmonize the diverse representations across two modalities.

For the text modality, CAA learns attentive semantic representations grounded on a modal-sharing token $R$. Specifically, the attribute text $S$ is regarded as the query to select distinct sharing tokens related to the attribute and generate the following output:
\begin{equation}
%{\bar{\mathcal{M}}}=Q_R\cdot K_{\hat{S}}^{\mathsf{T}}
\tilde{S}={\rm softmax}({\rm GMP}(Q_S\cdot K_{{R}}^{\mathsf{T}}))\cdot V_{{R}}
\label{eq:M'}
\end{equation}
where $Q_{\cdot}$, $K_{\cdot}$, and $V_{\cdot}$ are the linear mapping functions for the query, key, and value. $R$ is the modal-sharing token with a set of learnable parameters. $\rm GMP$ denotes the max-pooling operation posed on the token-level relevance between $S$ and $R$. This operation helps eliminate the influence of irrelevant noisy tokens that exhibit low relevance to modal-sharing tokens, producing a refined and compact concept-harmonized attribute token $\tilde{S}$.

Similarly, for the visual modality, within CAA, the visual feature $\bar{E}$ serves as the query, and $R$ serves as both the key and value:
\begin{equation}
\tilde{E}={\rm softmax}({\rm GMP}(Q_{\bar{E}}\cdot K_{{R}}^{\mathsf{T}}))\cdot V_{{R}}
\end{equation}
As a result, both image and attribute features are represented as combinations of the common modal-sharing token $R$. This explicit alignment ensures that the concepts of cross-modal information $\tilde{S}$ and $\tilde{E}$ are harmonized, which facilitates the subsequent prompt enhancement with a narrowed cross-modal gap.

\subsection{Semantic-Enhanced Prompt Distillation}
Based on the concept-harmonized outputs $\tilde{S}$ and $\tilde{E}$, these features help reveal a more comprehensive representation from both visual and semantic perspectives. Here, we apply the visual residual refinement to distill the semantic-enhanced prompt by predicting the details specific to attribute clues.

\subsubsection{Visual Residual Refinement Unit (VRRU)} 
Instead of relying on the commonly used attention mechanism, which has high computational complexity, VRRU adopts a simpler yet effective approach inspired by LLaVa~\cite{llava}. This unit utilizes a straightforward linear layer to establish a connection between the image embedding and the attribute embedding. Specifically, we first concatenate the outputs of CAA, \emph{i.e.}, $\tilde{S}$ and $\tilde{E}$, to fuse multiple modality features. Then, we employ a specialized linear prediction layer to estimate residual details. As a result, we have the following output:
\begin{equation}
Z = ZLinear \left( {\left[ {\tilde{S},\tilde{E}} \right]} \right)
\label{eq:linear}
\end{equation}
where $ZLinear$ function serves as a predictor with its weights initially set to zero~\cite{controlnet}, effectively eliminating harmful noise at the beginning of the training process.

\begin{table*}[t]
\centering 
\setlength\tabcolsep{3pt} 
\resizebox{7in}{!}{
\begin{tabular}{r|c|c|ccc|c|ccc|c|ccc}
\toprule
\multirow{3}{*}{\textbf{Methods}} &\multirow{3}{*}{\textbf{Venue}} &\multicolumn{4}{c|}{\textbf{CUB}}&\multicolumn{4}{c|}{\textbf{SUN}}&\multicolumn{4}{c}{\textbf{AwA2}}\\
\cmidrule{3-6}\cmidrule{7-10}\cmidrule{11-14} & &\multicolumn{1}{c|}{ZSL}&\multicolumn{3}{c|}{GZSL}&\multicolumn{1}{c|}{ZSL}&\multicolumn{3}{c|}{GZSL}&\multicolumn{1}{c|}{ZSL}&\multicolumn{3}{c}{GZSL}\\
\cmidrule{3-6}\cmidrule{7-10}\cmidrule{11-14} \textbf{}& &$acc$ &$U$ &$S$ &$H$ &$acc$&$U$ &$S$ &$H$ &$acc$&$U$ &$S$ &$H$ \\
\midrule
\multicolumn{1}{r@{}}{\textbf{Generative-based Methods}}\\
Composer~\cite{Huynh2020CompositionalZL} &NeurIPS'20&69.4&56.4&63.8&59.9&62.6&55.1 &22.0&31.4&71.5&62.1&77.3&68.8\\
GCM-CF~\cite{Yue2021CounterfactualZA}&CVPR'21 &--&61.0&59.7&60.3&--&47.9&37.8&42.2&--&60.4&75.1&67.0\\
SDGZSL~\cite{SDGZSL2021}&ICCV'21&75.5&59.9&66.4&63.0&--&--&--&--&72.1&64.6&73.6&68.8\\
CE-GZSL~\cite{CEGZSL2021}&CVPR'21&77.5&63.9&66.8&65.3&63.3&48.8&38.6&43.1&70.4&63.1&78.6&70.0\\
ICCE~\cite{ICCE2022} &CVPR'22&78.4 &67.3 &65.5 &66.4 &--&-- &-- &-- &72.7&65.3 &82.3 &72.8 \\ 
FREE~\cite{FREE2021} &ICCV'21&--&55.7&59.9&57.7&--&47.4&37.2&41.7&--&60.4&75.4&67.1\\
HSVA~\cite{HSVA2021} &NeurIPS'21&62.8&52.7&58.3&55.3&63.8&48.6&39.0&43.3&--&59.3&76.6&66.8\\
LBP~\cite{Lu2021ZeroAF} &TPAMI'21&61.9&42.7&71.6 &53.5 &63.2 &39.2 &36.9 &38.1 &--&--&--&--\\
f-VAEGAN+DSP~\cite{chen2023evolving}&ICML'23&62.8&62.5&73.1 &67.4&\underline{68.6}&\underline{57.7} &41.3&\underline{48.1}&71.6&63.7 &\textbf{88.8} &\underline{74.2}\\
SHIP$^\dagger$~\cite{ship} &ICCV'23&-- &55.3 &58.9 &57.1 &-- &-- &-- &-- &--&-- &--&-- \\\midrule
\multicolumn{1}{r@{}}{\textbf{Embedding-based Methods}}\\
APN~\cite{APN2020}&NeurIPS'20&72.0&65.3&69.3&67.2&61.6&41.9&34.0&37.6&68.4&57.1&72.4&63.9\\
DAZLE~\cite{DAZLE2020}&CVPR'20&66.0&56.7&59.6&58.1&59.4&52.3&24.3&33.2&67.9&60.3&75.7&67.1\\
DVBE~\cite{DVBE2020}&CVPR'20&--&53.2&60.2&56.5&--&45.0&37.2&40.7&--&63.6&70.8&67.0\\ 
GEM-ZSL~\cite{GEM2021}&CVPR'21&77.8&64.8&\underline{77.1}&70.4&62.8&38.1&35.7&36.9&67.3&64.8&77.5&70.6\\
DPPN~\cite{DPPN2021} &NeurIPS'21&77.8 &\underline{70.2} &\underline{77.1} &73.5 &61.5&47.9 &35.8 &41.0 &73.3 &63.1 &\underline{86.8} &73.1\\
GNDAN~\cite{Chen2022GNDANGN}&TNNLS'22&75.1&69.2&69.6&69.4&65.3&50.0&34.7&41.0&71.0&60.2&80.8&69.0\\
CLIP~\cite{clip} &ICML'21&-- &55.2 &54.8 &55.0 &-- &-- &-- &-- &--&-- &--&--  \\
CoOP$^\dagger$~\cite{coop} &IJCV'22&-- &49.2 &63.8 &55.6 &-- &-- &-- &-- &--&-- &--&--  \\
MSDN~\cite{MSDN2022}&CVPR'22&76.1&68.7&67.5&68.1&65.8&52.2&34.2&41.3&70.1&62.0&74.5&67.7\\
TransZero~\cite{TransZero} &AAAI'22&76.8&69.3&68.3&68.8&65.6&52.6&33.4&40.8&70.1&61.3&82.3&70.2\\
TransZero++~\cite{TransZero++} &TPAMI'22&78.3&67.5&73.6&70.4&67.6&48.6&37.8&42.5&72.6&64.6&82.7&72.5\\
DUET*$^\dagger$~\cite{chen2022duet} &AAAI'23&72.3 &62.9 &72.8 &67.5 &64.4 &45.7 &\underline{45.8} &45.8 &69.9 &63.7 &84.7 &72.7 \\
I2MVFormer*~\cite{I2MVFormer} &CVPR'23&42.1&32.4 &63.1 &42.8 &--&-- &-- &-- &\underline{73.6} &\underline{66.6} &82.9&73.8\\
ZSLViT*~\cite{ZSLViT} &CVPR'24&\underline{78.9} &69.4 &\textbf{78.2} &\underline{73.6} &68.3&45.9 &\textbf{ 48.3} &47.3 &70.2 &66.1 &84.6 &\underline{74.2}\\
\midrule
{\textbf{AENet*}}{~\textbf{(Ours)}} &--&\textbf{ 80.3} &\textbf{ 73.1}&76.4&\textbf{ 74.7} &\textbf{ 70.4} &\textbf{ 58.6}&45.2 &\textbf{ 51.0} &\textbf{ 75.2} &\textbf{ 70.3} & 80.1&\textbf{ 74.9} \\
\bottomrule	
\end{tabular} 
}
\caption{Results ~($\%$) of the state-of-the-art ZSL and GZSL models on CUB, SUN, and AwA2, including both generative and embedding-based methods. The symbol ``$*$'' denotes ViT-based methods. The symbol ``$\dagger$'' indicates the prompt-based methods, with results reported in \cite{ZSLViT}.}
\label{table:sota}
\end{table*}

\subsubsection{Enhancing Prompt for Visual Enrichment} 
To ensure the predicted residual $Z$ carries meaningful and semantically consistent information, we employ an attribute-guided optimization strategy by aligning $Z$ with category attribute prototypes, denoted as $a_{y}$, through a consistency loss $\mathcal{L}_{cons}$:
\begin{equation}
\mathcal{L}_{cons}=\|Z-{\rm MLP}(a_y)\|
\label{eq:L_sem}
\end{equation} 
where ${\rm MLP}$ refers to a multi-layer perceptron that projects the attribute prototypes into the same space as $Z$. This consistency loss function encourages $Z$ to capture attribute-specific features that are crucial for ZSL tasks.

Then, $Z$ is utilized to enhance the visual prompt by merging it back to the original prompt embedding via a skip connection:
\begin{equation}
\tilde P =  {\left[ {\bar{p}_1+Z,\bar{p}_2+Z,..., \bar{p}_T+Z} \right]}
\label{eq:skip}
\end{equation} 
$Z$ is added to each token ($\bar{p}_i$) of the original prompt, which can augment each prompt token with the attribute-aligned information from $Z$. The generated semantic-enhanced prompt $\tilde P$ incorporates attribute-related details derived from the multimodal information, allowing it to capture fine-grained semantics for distinguishing between similar classes in zero-shot scenarios. Finally, we integrate the enhanced prompt $\tilde P$ with the visual token $\bar E$ to enrich the visual representation $f(x)$ for $x$:
\begin{equation}
f(x)=\left[ \tilde P, \bar E\right]
\end{equation}

Together with the enhanced prompt, AENet is expected to achieve improved comprehensive transferable knowledge discovery, containing complementary information that may be overlooked by conventional prompt-based visual content.

\subsection{Model Optimization and Inference}
\subsubsection{Optimization}
The overall objective loss function of AENet is formulated as follows:
\begin{equation}
\mathcal{L}=\mathcal{L}_{cls} +\lambda_{cons}\mathcal{L}_{cons}+ \lambda_{deb}\mathcal{L}_{deb}
\label{eq:loss}
\end{equation}
where $\mathcal{L}_{cls}$ is the classification loss (Eq. (\ref{eq:cls})). $\lambda_{cons}$ and $\lambda_{deb}$ are the hyper-parameters controlling the weights of semantic consistency loss $\mathcal{L}_{cons}$ and the debiasing loss $\mathcal{L}_{deb}$, respectively.
In addition, as expressed in Eq. (\ref{eq:loss}), we also apply a debiasing loss $\mathcal{L}_{deb}$ to mitigate the seen-unseen bias following \cite{PSVMA2023,POPRNet}. It aims to balance the score dependency in the seen-unseen domain, pursuing the distribution consistency concerning both mean and variance:
\begin{equation}
{{\cal L}_{deb}} = \|{\alpha_s} - {\alpha_u}\|_2^2 + \|{\beta_s} - {\beta_u}\|_2^2
%{{\cal L}_{deb}} = max\{0,{\alpha_s} - {\alpha_u}\} + max\{0,{\beta_s} - {\beta_u}\}
\end{equation}
where $\alpha_s$ and $\beta_s$ represent the mean and variance, respectively, of the seen prediction score $\left\langle {\mathcal{M}(f(x)),{a_{\hat y(\hat y\in \mathcal{Y}^{s})}}} \right\rangle$. Similarly, $\alpha_u$ and $\beta_u$ denote the mean and variance, respectively, of the unseen prediction score $\left\langle {\mathcal{M}(f(x)),{a_{\hat y(\hat y\in \mathcal{Y}^{u})}}} \right\rangle$.

\subsubsection{Inference} 
During training, the model merely learns about the knowledge of seen categories, while unseen categories are inferred at testing time:
\begin{equation}
\tilde{y}=\arg \max_{\hat{y} \in  \mathcal{Y}^u}\left(\left\langle {\mathcal{M}(f(x)),{a_{\hat y}}} \right\rangle\right)
\end{equation}
In the GZSL setting, both seen and unseen categories are encompassed.
To jointly define the category, calibrated stacking (CS) \cite{2016cs} is applied:
\begin{equation}
\tilde{y}=\arg \max_{\hat{y} \in \mathcal{Y}}\left(\left\langle {\mathcal{M}(f(x)),{a_{\hat y}}} \right\rangle-\gamma\mathbb{I}_{\left[\hat{y} \in \mathcal{Y}^s\right]}\right)
\label{eq:pre}
\end{equation}
where $\mathbb{I}_{\mathcal{Y}^{S}}(\cdot)$ represents an indicator function, yielding a result of 1 when $\hat{y} \in \mathcal{Y^S}$ and 0 otherwise. The calibrated factor $\gamma$ is employed to trade off the calibration degree on seen categories and determine the category $\tilde{y}$ of an input visual sample $x$.

\begin{table*}[ht]
\centering 
\resizebox{7in}{!}{
\begin{tabular}{c|c|ccc|c|ccc|c|ccc}
\toprule
\multirow{3}{*}{\textbf{Methods}}&\multicolumn{4}{c|}{\textbf{CUB}}&\multicolumn{4}{c|}{\textbf{SUN}}&\multicolumn{4}{c}{\textbf{AwA2}}\\
\cmidrule{2-5}\cmidrule{6-9}\cmidrule{10-13} &\multicolumn{1}{c|}{ZSL}&\multicolumn{3}{c|}{GZSL}&\multicolumn{1}{c|}{ZSL}&\multicolumn{3}{c|}{GZSL}&\multicolumn{1}{c|}{ZSL}&\multicolumn{3}{c}{GZSL}\\
\cmidrule{2-5}\cmidrule{6-9}\cmidrule{10-13} &$acc$ &$U$ &$S$ &$H$ &$acc$&$U$ &$S$ &$H$ &$acc$&$U$ &$S$ &$H$ \\
\midrule
AENet w/o prompt $P$ & 70.2& 67.0&69.1&68.0 &64.5 &53.8&25.8 &34.8 &70.7 &61.0 & 88.2&72.1 \\
AENet w/o residual details $Z$& 75.8& 72.3&72.7&72.5 &68.0 &58.1&38.5 &46.2 &73.2 &65.5 & 85.1&74.0 \\
AENet w/o concept-aware attention &78.8 &75.6 &72.1&73.8 &68.8 & 43.6&54.9 &48.6 &73.7 &66.8 &81.8 &73.5 \\
AENet (w/ all)& 80.3& 73.1&76.4&74.7 &70.4 &58.6&45.2 &51.0 &75.2&70.3 & 80.1&74.9 \\
\bottomrule	
\end{tabular}}
\caption{Ablation study of AENet under the ZSL and GZSL settings on CUB, SUN, and AwA2 datasets.}
\label{table:ablation}
\end{table*}
\begin{table}[ht]
\centering 
\begin{tabular}{c|cc|cc}
\toprule
\multirow{2}{*}{\textbf{Methods}}&\multicolumn{2}{c|}{\textbf{CUB}}&\multicolumn{2}{c}{\textbf{AwA2}}\\
&$acc$ &$H$ &$acc$&$H$ \\
\midrule
Linear + Skip & 79.8 & 74.5 &74.1 &74.5\\
MLP + Skip & 79.5&74.0 &74.8 &74.3\\
ZLinear + Gated Skip & 79.5&74.4 &73.5 &74.4\\
ZLinear + Skip & 80.3&74.7&75.2 &74.9 \\
\bottomrule	
\end{tabular}
\caption{Ablation study of different implementations for semantic-enhanced prompt distillation.}
\label{table:ablation2}
\end{table}

\section{Experiments}
\subsection{Experimental Settings}
\noindent
\subsubsection{Datasets}
We conduct experiments on three standard benchmark datasets: Caltech-UCSD Birds-200-2011 (CUB) \cite{DatasetCUB}, SUN Attribute (SUN) \cite{DatasetSUN}, Animals with Attributes2 (AwA2) \cite{DatasetAWA2}. The categorization into seen and unseen categories follows the Proposed Split (PS) \cite{DatasetAWA2}. The CUB dataset consists of 11,788 images illustrating 200 bird classes, with a split of 150/50 for seen/unseen classes and characterized by 312 attributes. SUN is a vast scene dataset that contains 14,340 images spanning 717 classes, divided into seen/unseen classes at 645/72 and annotated with 102 attributes. AwA2 contains 37,322 images of 50 animal classes, with a 40/10 split for seen/unseen classes, and is described by 85 attributes.

\noindent
\subsubsection{Evaluation Metrics}
We evaluate top-1 accuracy in both the ZSL and GZSL settings. For ZSL, we calculate accuracy solely on unseen classes, denoted as $acc$. In the GZSL setting, following~\cite{DatasetAWA2}, we employ the harmonic mean (as $H = 2 \times S \times U / (S + U)$) to measure performance, where $S$ and $U$ represent the top-1 accuracy of the seen and unseen classes, respectively.

\subsubsection{Implementation Details}
we apply the ViT-Base model \cite{ViT2020} as visual feature extractor. The input image resolution is $224 \times 224$, with a patch size of $16 \times 16$. Our framework is implemented using PyTorch and executed on an NVIDIA GeForce RTX 3090 GPU.

\subsection{Comparison with State-of-the-Art Methods}
We evaluate our AENet and compare it with recently state-of-the-art methods. The results are presented in Table \ref{table:sota}. 

\subsubsection{Results of ZSL}
For conventional ZSL, our method demonstrates significant $acc$ improvements of 1.4\%, 1.8\%, and 1.6\% on CUB, SUN, and AwA2 datasets, surpassing the previous state-of-the-art methods \cite{ZSLViT,chen2023evolving,I2MVFormer}. AENet facilitates more comprehensive knowledge transference for unseen classification, thus achieving state-of-the-art accuracy performance of 80.3\%, 70.4\%, and 75.2\% on CUB, SUN, and AwA2, respectively. Compared to recent methods \cite{GEM2021,TransZero,TransZero++,ZSLViT} that refine visual features obtained from pre-trained visual encoders using attention mechanisms, AENet achieves substantial improvements in $acc$, with gains exceeding 1.4\%, 2.1\%, and 2.6\% on CUB, SUN, and AwA2, respectively. Furthermore, our approach demonstrates superior performance compared to the prompt-based ZSL method \cite{chen2022duet}, yielding obvious improvements of 8.0\%, 6.0\%, and 5.3\% on the CUB, SUN, and AwA2 datasets, respectively.

\subsubsection{Results of GZSL}
Table \ref{table:sota} also reports the results of various methods in the GZSL setting. Results show that AENet can achieve state-of-the-art $H$ performance across all datasets, \emph{i.e.}, 74.7\%, 51.0\%, and 74.9\% on CUB, SUN, and AwA2, respectively. Similar to ZSL, our AENet also significantly outperforms other prompt-based methods \cite{coop,ship,chen2022duet} with substantial margins of 7.2\%, 5.2\%, and 2.2\% on CUB, SUN, and AwA2 datasets, respectively. These results demonstrate that AENet significantly enhances cross-domain transferability by enriching the visual representations.

\begin{figure}[t]
\begin{center}
\includegraphics[scale=0.6]{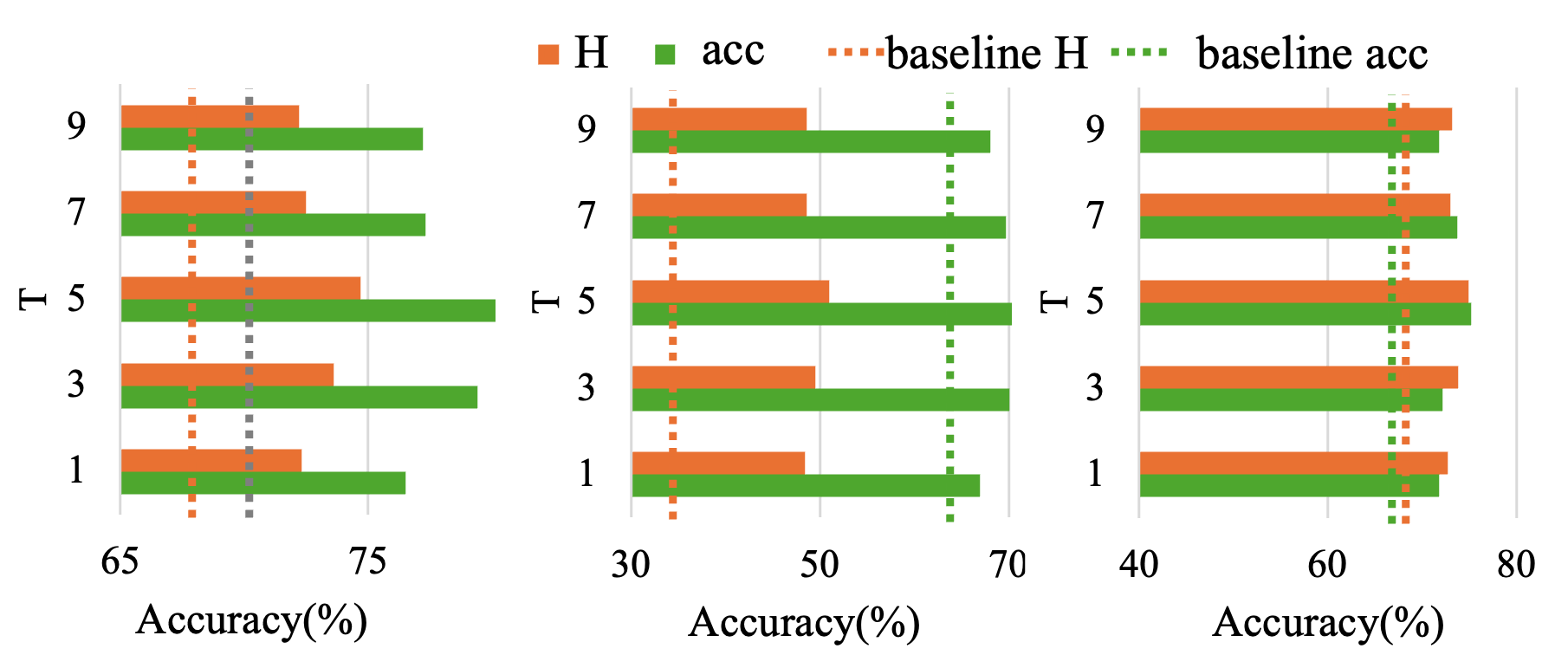}
\caption{Effect of the length ($T$) of $P$ on CUB, SUN and AwA2 datasets.}
\label{fig:tnum}
\end{center}
\end{figure}

\subsection{Ablation Study}

\subsubsection{Effect of components in AENet} As shown in Table \ref{table:ablation}, we evaluate key components in AENet, \emph{i.e.}, the learnable prompt $P$, semantic-related details $Z$ for the semantic-enhanced prompt, and concept-aware attention for the concept-harmonized token generation. When the original learnable prompt is removed, the model can be regarded as the baseline that directly uses the visual features extracted from vanilla ViT and projects them into semantic space for category inference. AENet outperforms the baseline with large $acc$/$H$ margins of 20.1\%/6.7\%, 5.9\%/6.2\%, and 4.5\%/2.8\% on CUB, SUN, and AwA2 datasets, respectively. The performance of AENet degrades when residual details are removed, as evidenced by decreases in both ZSL and GZSL across multiple datasets: specifically, a reduction in $acc$/$H$ of 4.5\%/2.2\% on CUB, 2.4\%/4.8\% on SUN, and 2.0\%/0.9\% on AwA2, compared to the full model implementation. Additionally, concept-aware attention provides concept-harmonized visual and semantic representations, resulting in $acc$/$H$ improvements of 1.5\%/0.9\%, 1.6\%/2.4\%, 1.5\%/1.4\% on CUB, SUN, and AwA2 datasets, respectively. Additionally, we implement variants by replacing ZLinear in Eq. (\ref{eq:linear}) with MLP and a linear layer, and the skip connection in Eq. (\ref{eq:skip}) with a gated skip. As demonstrated in Table \ref{table:ablation2}, these more complex alternatives do not offer performance improvements as the harmonized tokens are concise and informative. Moreover, ZLinear consistently achieves better performance compared to the linear.

\begin{figure}[t]
\begin{center}
\includegraphics[scale=0.54]{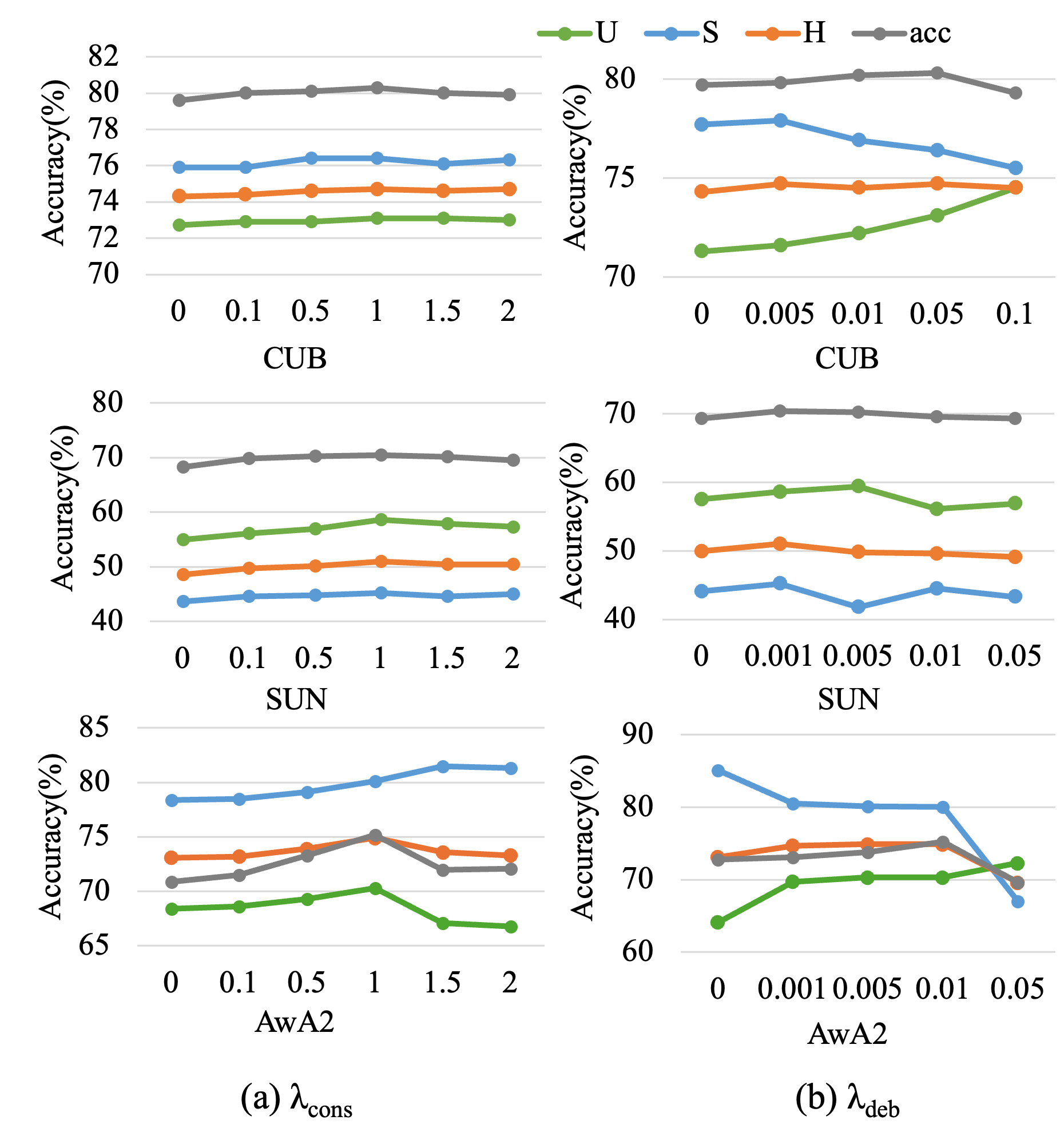}
\caption{Effect of loss hyper-parameters.}
\label{fig:loss}
\end{center}
\end{figure}

\begin{figure}[t]
\begin{center}
\includegraphics[scale=0.52]{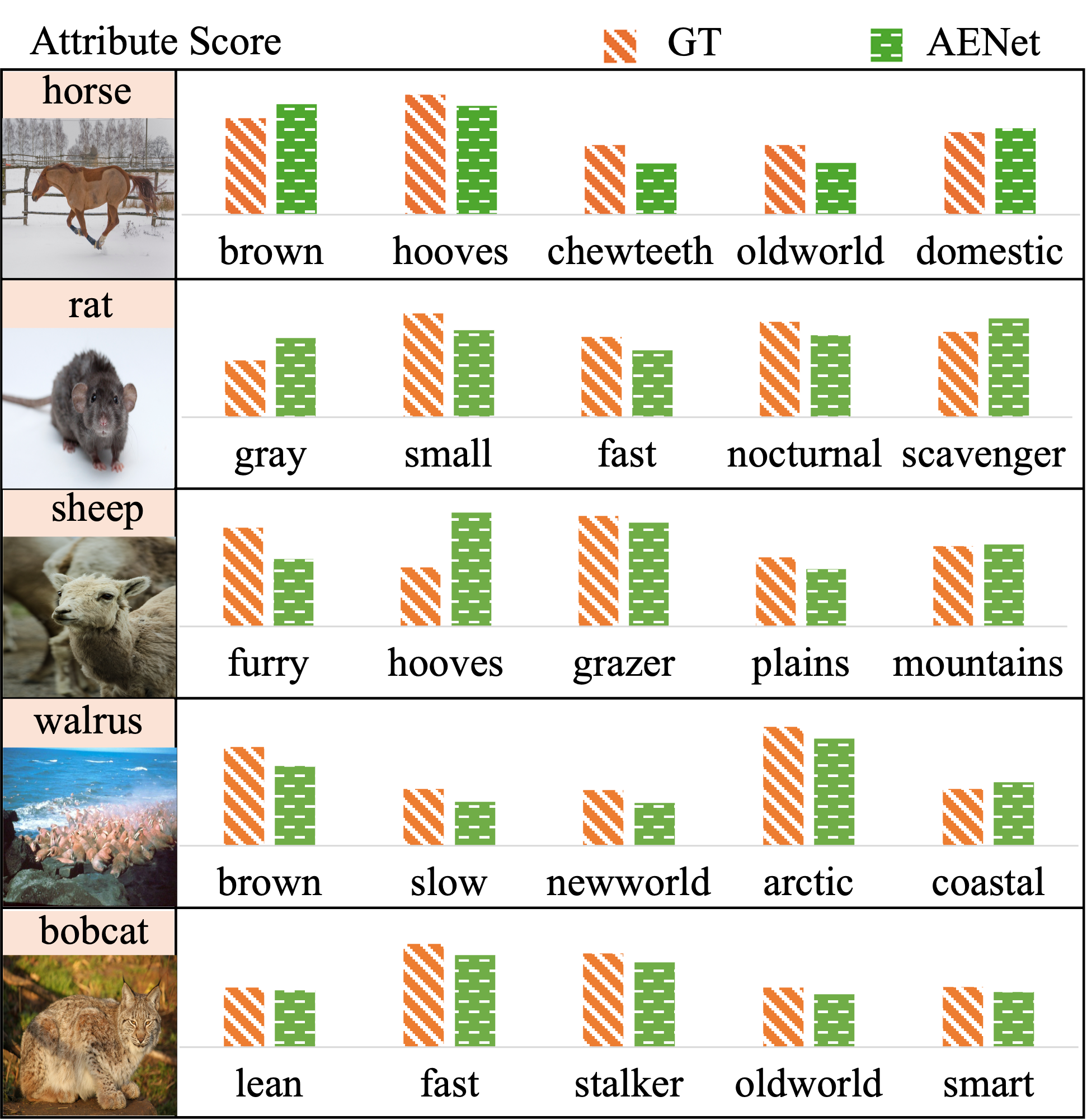}
\caption{Attribute prediction for visualization.}
\label{fig:attribute}
\end{center}
\end{figure}

\begin{figure}[t]
\begin{center}
\includegraphics[scale=0.45]{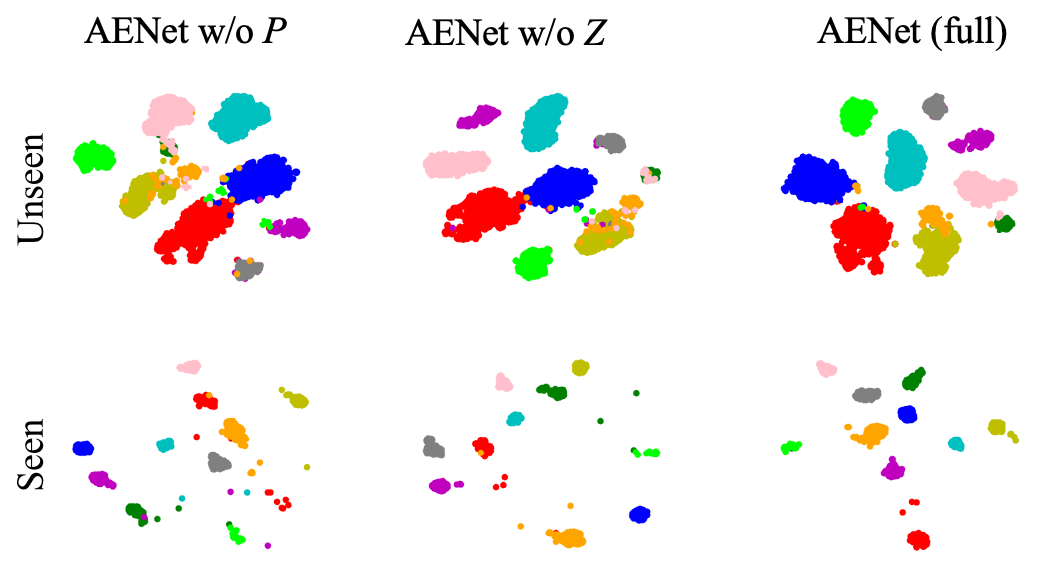}
\caption{t-SNE visualizations of visual features for seen classes and unseen classes. The 10 colors denote different classes randomly selected from the AwA2 dataset.}
\label{fig:tsne}
\end{center}
\end{figure}

\subsubsection{Effect of $T$} $T$ is the length of the learnable visual prompt $P$. Here, we sweep prompt length $T \in \{ 1, 3, 5, 7, 9\}$ to investigate the effect of the prompts $P$ on classification performance. Figure \ref{fig:tnum} shows the values of $acc$ and $H$ as $T$ varies. We can observe that the best performance is achieved when $T$ is approximately 5. Notably, even with a single prompt, AENet demonstrates significant performance improvements over the baseline (1$st$ row in Table \ref{table:ablation}). Based on these findings, we set $T$ = 5 for CUB, SUN, and AwA2 datasets.

\subsubsection{Impact of $\lambda_{cons}$ and $\lambda_{deb}$} In this work, we combine the attribute consistency loss $\mathcal{L}_{cons}$ and $\mathcal{L}_{deb}$ and debiasing with the balance hyper-parameters of $\lambda_{cons}$ and $\lambda_{deb}$ respectively to train the model. Here, we study the impact of $\lambda_{cons}$ and $\lambda_{deb}$ as shown in Figure \ref{fig:loss}. As $\lambda_{cons}$ rises from 0.0 to 2.0, \emph{i.e.}, the attribute-supervision $\mathcal{L}_{cons}$ is introduced into AENet for the semantic prediction, $H$ increases on all datasets. The best $H$ is obtained when $\lambda_{cons} = 1.0$. This demonstrates the effectiveness of semantic consistency in predicting semantically relevant details, closely aligning with category prototypes to focus on attribute-specific features. When $\lambda_{cons} > 1.0$, $H$ decreases. Thus, we set $\lambda_{cons} = 1.0$ for optimal results. Besides, increasing $\lambda_{deb}$ leads to more consistent distribution between seen and unseen predictions, improving unseen accuracy $U$ and overall $H$ performance.

\subsection{Visualization}
\subsubsection{Attribute Prediction}
To intuitively validate the capability of AENet for capturing enhanced semantic information necessary for unseen classes, we input unseen images into AENet to predict the attribute scores ($\mathcal{M}(f(x))$). As shown in Figure \ref{fig:attribute}, we can observe that AENet demonstrates strong performance in predicting attributes across various animal species in the AwA2 dataset. For example, for the horse, AENet closely matches the ground truth (GT) on ``brown'', ``hooves'', and ``domestic'', with slight discrepancies in ``chewteeth'' and ``oldworld''. The rat predicted scores of the rat are remarkably accurate across all attributes, showcasing precision for this species. These results illustrate AENet's robust capability in capturing enhanced semantic information across diverse species, which is crucial for generalizing to unseen classes in ZSL tasks.

\subsubsection{T-SNE Visualization}
As shown in Figure \ref{fig:tsne}, we present the t-SNE visualization of visual features for seen and unseen classes. Compared to full AENet, visual features extracted from the AENet w/o $P$ lack distinctiveness within certain classes. This intuitively suggests that the prompt plays a crucial role in enabling the application of pre-trained ViT to downstream ZSL tasks by obtaining high-quality features for seen classes. Furthermore, compared to AENet w/o $Z$, the visual features learned by our full AENet showcase desirable distinguishability with higher inter-class discrepancy and clearer decision boundaries. 

\section{Conclusion}
This paper introduces AENet, which leverages prompt learning and addresses the challenge of over-emphasizing primary visual features from seen domains. AENet comprises two key steps. Firstly, the concept-harmonized tokens are explored for visual and attribute modalities based on the modal-sharing token with consistent concepts. Secondly, by investigating the semantic-related details, the VRRU is proposed to distill the semantic-enhanced prompt integrated with primary visual features to attend to semantic-related information for visual enrichment. Consequently, AENet can achieve superior visual-semantic alignment for generalizing to unseen classes. Extensive experiments across three benchmark datasets show the superiority of AENet.

\section{Acknowledgments}
%\texttt{\textbackslash section*\{Acknowledgements\}}
This work was supported in part by Fundamental Research Funds for the Central Universities (2024JBZY001, JZ2024HGTB0255, 2024XKRC082), National Natural Science Foundation of China (No. 62331003, 62120106009, 62302141, 62476021, 72434005, 92470203), Joint Funds of the National Natural Science Foundation of China under Grant U23A20314, Beijing Natural Science Foundation (L223022, L242022), Natural Science Foundation of Hebei Province (F2024105029), the Chinese Association for Artificial Intelligence (CAAI)-Compute Architecture for Neural Networks (CANN) Open Fund, developed on OpenI Community.

\bibliography{aaai25}

\end{document}